\documentclass{article}
\usepackage{amsmath, bm, graphicx, mlspconf, glossaries}
\usepackage{blindtext}
\usepackage{hyperref}
\usepackage{afterpage}
\usepackage{multicol}
\usepackage{float}
\usepackage{epstopdf}
\usepackage{xcolor}

\makeatletter
\renewcommand{\section}{\@startsection{section}{1}{\z@}%
  {-2.0ex \@plus -0.5ex \@minus -.2ex}%
  {1.50ex \@plus.2ex \@minus-.2ex}%
  {\normalfont\large\bfseries\centering}}
  
\renewcommand{\subsection}{\@startsection{subsection}{2}{\z@}%
  {-1.50ex\@plus -1ex \@minus -.2ex}%
  {0.50ex \@plus .2ex}%
  {\normalfont\normalsize\bfseries}}
\makeatother

\copyrightnotice{979-8-3503-2411-2/23/\$31.00 {\copyright}2023 IEEE}

\toappear{2023 IEEE International Workshop on Machine Learning for Signal Processing, Sept.\ 17--20, 2023, Rome, Italy}

\title{Concept-based explainability for an EEG transformer model}

\name{%
    \fontsize{10.5pt}{11pt}\selectfont
    \begin{tabular}{c}
    Anders Gjølbye Madsen$^{\star\dagger}$ \qquad William Theodor Lehn-Schiøler$^{\star\dagger}$ \\
    Áshildur Jónsdóttir$^{\star}$ \qquad Bergdís Arnardóttir$^{\star}$ \qquad Lars Kai Hansen$^{\star}$
    \end{tabular}
    \thanks{This work is supported by The Pioneer Centre for AI, DNRF grant number P1, The Novo Nordisk Foundation grant NNF22OC0076907 "Cognitive spaces - Next generation explainability", and travel grants from The Danish Data Science Academy awarded to AGM and WLS.}
    \vspace{-3pt}
}
\address{%
    \small
    \begin{tabular}{c}
    \begin{tabular}{@{}c@{}}
    $^{\star}$Technical University of Denmark \\
    Department of Applied Mathematics and Computer Science \\
    2800 Kgs. Lyngby, Denmark
    \end{tabular}
    \qquad \qquad
    \begin{tabular}{@{}c@{}}
    $^{\dagger}$BrainCapture \\
    2800 Kgs. Lyngby, Denmark
    \end{tabular}
    \end{tabular}
    \vspace{-11pt}
}

\begin{document}
\ninept

\maketitle

\begin{abstract}
Deep learning models are complex due to their size, structure, and inherent randomness in training procedures. Additional complexity arises from the selection of datasets and inductive biases. Addressing these challenges for explainability, Kim et al. (2018) introduced Concept Activation Vectors (CAVs), which aim to understand deep models' internal states in terms of human-aligned concepts. These concepts correspond to directions in latent space, identified using linear discriminants. Although this method was first applied to image classification, it was later adapted to other domains, including natural language processing.
In this work, we attempt to apply the method to electroencephalogram (EEG) data for explainability in Kostas et al.'s BENDR (2021), a large-scale transformer model. A crucial part of this endeavour involves defining the explanatory concepts and selecting relevant datasets to ground concepts in the latent space. Our focus is on two mechanisms for EEG concept formation: the use of externally labelled EEG datasets, and the application of anatomically defined concepts. The former approach is a straightforward generalization of methods used in image classification, while the latter is novel and specific to EEG.
We present evidence that both approaches to concept formation yield valuable insights into the representations learned by deep EEG models.
\end{abstract}

\begin{keywords}
Explainable AI, EEG Concepts, TCAV, BENDR
\end{keywords}

\section{Introduction} \label{sec:intro}
\begin{figure*}[htb]
    \centering
    \includegraphics[width=0.9\textwidth]{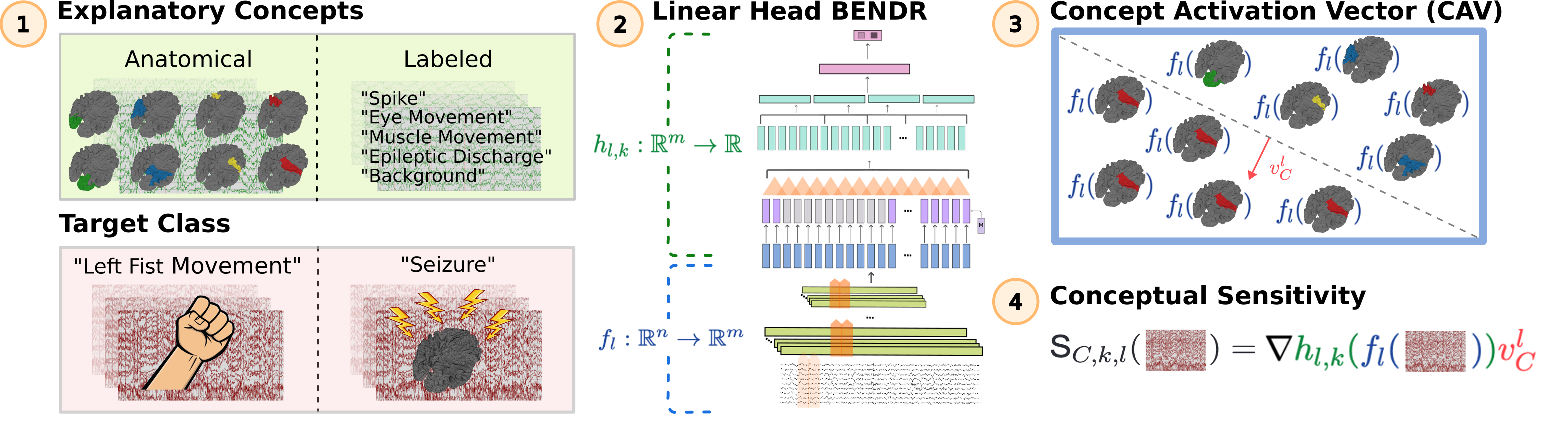}
    \vspace{-3pt}
    \caption{An overview of using the TCAV method for EEG classification tasks with the Linear Head BENDR model: (1) Explanatory concepts are defined as either event-based EEG labels or frequency-based cortical activity, (2) Layer activations are extracted from a fine-tuned Linear Head BENDR, (3) Concept Activation Vectors (CAV) are defined as the normal vector to the hyperplane separating layer activations for concept data from those of random examples, and (4) The sensitivity of class data for a specific bottleneck of a concept is defined as the directional derivative in the direction of the respective CAV.}
    \label{fig:figure_1}
    \vspace{-11pt}
\end{figure*}

We investigate representations of electroencephalogram (EEG) data obtained by self-supervised learning methods. Self-supervision is motivated by the lack of labeling in large-scale EEG datasets as labeling is both time-consuming and requires highly specialised EEG expertise.
Self-supervised models, such as BERT-inspired Neural Data Representations (BENDR) \cite{BENDR}, have the potential to overcome this challenge by learning informative representations from raw, unlabeled data. Such models can subsequently be fine-tuned for downstream classification tasks. We apply the Testing Concept Activation Vectors (TCAV) approach of Kim et al. \cite{kim2018interpretability}, an interpretability method introduced in 2018, to BENDR-based models, to provide insights into their structure and decision-making processes. See Figure \ref{fig:figure_1} for a conceptual overview.  A better understanding of EEG transformer models using TCAV could support the use of these models as diagnostic support tools for identifying EEG abnormalities, such as seizures. However, the question that arises is, what constitutes human-friendly concepts in this context? To address this, we present the following scientific contributions:
%
\begin{itemize}
\itemsep 0pt
\parsep 0pt
    \item The first TCAV workflows for EEG data, proposing concepts based on human-annotated data as well concepts defined by human anatomy and EEG frequency ranges.
    \item Sanity checks for TCAV to ensure valid explanations in simple EEG settings.
    \item Two practical applications: seizure prediction and brain-computer interfacing.
\end{itemize}

All code used in this research, along with references to the datasets, have been made publicly accessible for validation and replication\footnote{\url{https://github.com/AndersGMadsen/TCAV-BENDR}}.

%
\section{Theory} \label{sec:theory}
\subsection{BERT-inspired Neural Data Representations}
\label{ssec:bendr}
BENDR \cite{BENDR} is inspired by language modeling techniques that have found success also outside text analysis, in self-supervised end-to-end speech recognition and image recognition. It aims to develop EEG models for better brain-computer interface (BCI) classification, diagnosis support, and other EEG-based analyses. Importantly, the approach being based on self-supervision can learn from any EEG data using only unlabeled data.
The main goal of BENDR is to create self-supervised representations with minimal robust to context boundaries  like datasets and  human subjects. The approach is expected to be transferable to future unseen EEG datasets recorded from unseen subjects, different hardware, and different tasks. It can be used as-is or fine-tuned for various downstream EEG classification tasks.

\quad The architecture is based on \verb|wav2vec 2.0| \cite{wave2vec} developed for speech processing and  consists of two stages. The first stage takes raw data, and down-samples it using a stack of short-receptive field 1D convolutions, resulting in a sequence of vectors called BENDR. The second stage uses a transformer encoder \cite{vaswani2017attention} to map BENDR to a new sequence related to the target task. Down-sampling is achieved through strides, and the transformer follows the standard implementation with some modifications. The entire sequence is then classified, with a fixed token implemented as the first input for downstream tasks \cite{BERT}.
BENDR differs from the speech-specific architecture  in two ways: (1) BENDR is not quantized for pre-training targets, and (2) it has many incoming channels, unlike \verb|wav2vec 2.0| which uses quantization and is based on a single channel of raw audio. The 1D convolutions are preserved in BENDR, to reduce complexity. We note that BENDR down-samples at a lower factor than \verb|wav2vec 2.0|, here resulting in an effective sampling rate of $\approx 2.67$ Hz equivalent to a feature window of $\approx 375$ ms.

\subsection{Linear Head BENDR}
\begin{figure}[htb]
\begin{minipage}[b]{
\linewidth}
  \centering
  \centerline{\includegraphics[width=0.9\textwidth
  ]{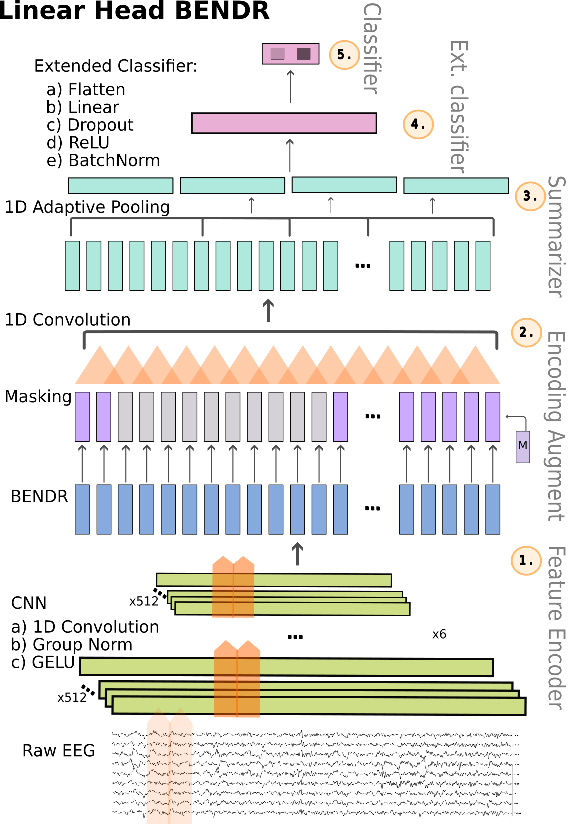}}
    \caption{The Linear Head BENDR (LHB) model architecture illustrated. The model consists of (1) Feature encoder of six confrontational blocks, (2) Encoding augment comprised of masking and convolutional contextualizer, (3) Summarizer using Adaptive Average Pooling, (4) Extended Classifier for dimensionality reduction, and (5) Classifier.}
    \label{fig:linear_head_bendr}
    \vspace{-11pt}
\end{minipage}
\end{figure}

For downstream fine-tuning, we use a version where the pre-trained transformer modules are ignored, such that the pre-trained convolutional BENDR stage is used as representation, see \cite{BENDR}. A consistent-length representation is created by dividing the BENDRs into four contiguous sub-sequences, averaging each sub-sequence, and concatenating them. A new linear layer with softmax activation is added to classify the downstream targets based on this concatenated vector of averaged BENDR. We call this the Linear Head BENDR (LHB) model and the structure is illustrated in Figure \ref{fig:linear_head_bendr}. 

The final LHB architecture consists of the following components:
\begin{enumerate}
    \item \textbf{Feature encoder:} Fine-tunes the pre-trained parameters and uses six convolution blocks, each containing a temporal convolution, group normalization, and a GELU activation function to produce a BENDR of length 512.
    \item \textbf{Encoding augment:} Involves masking and contextualizing the BENDR, with 10\% of the BENDR masked and 10\% of the channels dropped, while relative positional embeddings from the pre-trained task are added to the BENDR and further preprocessed.
    \item \textbf{Summarizer:} Applies adaptive average pooling to create four contiguous sub-sequences, averaging each sub-sequence to ensure the model's independence from the input length of EEG recordings.
    \item \textbf{Extended classifier:} Flattens the four sub-sequences, passes them through a fully connected layer to reduce their dimension, applies a dropout layer, uses a ReLU activation function, and normalizes the output using batch normalization.
    \item \textbf{Classifier:} Consists of a linear layer with a softmax activation function, which performs the classification task.
\end{enumerate}

%
\subsection{Testing with Concept Activation Vectors (TCAV)}
\label{ssec:tcav}
Testing with Concept Activation Vectors (TCAV) is a technique used to quantify the degree to which layers of neural networks align with human-defined concepts \cite{kim2018interpretability}. The method is general in the sense that it is not confined to the particular structure of the network nor to the data type. In its essence, TCAV can be broken down into five steps

\quad First, the process involves defining human-aligned concepts and representing them in the data. Alongside these, data from the target class must also be present for evaluation purposes. Furthermore, to establish the directions of the concept activation vector in the latent space, it is necessary to have a collection of concept-negative or random examples.

\quad Second, the layer activations of the concept input and the random input, respectively, are collected and separated by training 
a binary linear classifier. Then, the concept activation vector, $v_c^l$ is defined as the normal vector to the hyperplane that separates the two classes (concept vs. random).

\quad Third, for a layer $l$ in the network, the directional derivatives for the target class $k$ along the
learned activation vector for concept $C$ is used to calculate how sensitive the prediction of the network is to changes in the input data in the direction of $C$. We can quantify the sensitivity by
\begin{equation}
    S_{C,k,l}(\bm{x}) = \nabla h_{l,k} (f_l(\bm{x}))\cdot \bm{v}_C^l,
\end{equation}
where $h_{l,k}$ is defined as the function that maps activations in layer $l$ through the remaining network and predicts class $k$.

\quad Fourth, computing the sensitivity for several target examples, $\bm{x} \in X_k$, the TCAV score is defined as the ratio of examples that have positive sensitivity, i.e.,
\begin{equation}
    \vspace{-3pt}
    \text{TCAV}_{C,k,l} = \frac{|\{\bm{x} \in X_k : S_{C,k,l}(\bm{x}) > 0 \}|}{|X_k|}.
\end{equation}
In this way, concept activation vectors that are positively aligned with target activations have a TCAV score close to 1 and concept activation vectors that are negatively aligned with target activations have a TCAV score close to 0.

\quad Fifth and final, collecting samples of TCAV scores over several training runs, a suitable statistical test is used to assess the statistical significance of concept activation vectors aligning with the activation of target examples. The null hypothesis of the test is that half of the examples have positive sensitivity and the other half have negative or zero sensitivity, i.e.,
\begin{equation}
    H_0: \text{TCAV}_{C,k,l} = 0.5.
\end{equation}
Concepts $C$ for which the null hypothesis is rejected thus relate to the target class prediction, and may bring positive or negative evidence for the given target $k$.
\subsection{Source localization}
\label{ssec:sourcelocalisation}
Source localization for EEG data involves mapping electrical signals recorded on the scalp surface to corresponding regions on the cortical surface of the brain. This process uses a head model and the EEG data collected from electrodes placed on the scalp. The reconstruction is a grid of dipolar sources. The solution to this ill-posed problem is called the lead field and there exist many different ways to obtain this solution. In this work, we use the exact low-resolution electromagnetic tomography (eLORETA) method implemented in the MNE library \cite{doi:10.1098/rsta.2011.0081}.

\quad The eLORETA approach presupposes that the EEG measurements of the electric field present on the scalp reflect dipolar sources located in the cerebral cortex. These are conceptually modeled as a three-dimensional distribution of dipoles. The spatial resolution of eLORETA is relatively coarse, which can make pinpointing exact cortical sources challenging. However, for our purpose of estimating aggregated source activity over broadly defined brain regions, such reduced resolution is not an issue.


\section{Methods} \label{sec:methods}
\subsection{Data} \label{ssec:data}

EEG is a non-invasive technique to record the brain's electrical activity. EEG data in this paper refers to these measurements, used often in research and healthcare to identify neurological conditions. In this work, we use five publicly accessible datasets, namely TUH EEG Corpus \cite{tuheeg}, TUH EEG Artifact (TUAR) Corpus, TUH EEG Events (TUEV) Corpus, TUH EEG Seizure (TUSZ) Corpus \cite{tuhz} and the EEG Motor Movement/Imagery (MMIDB) Dataset \cite{mmidb}.

\quad The TUH EEG Corpus contains 69,652 clinical and unlabeled EEG recordings obtained from Temple University Hospital (TUH). The TUH EEG Artifact Corpus, a labeled subset of the TUH EEG Corpus, includes annotations for five distinct artifacts including eye movement artifact (\emph{eyem}). The TUEV is a subset of the TUH EEG Corpus and includes annotations of event-based EEG segments. There are numerous categories, but we primarily focus on five key classes: (1) technical artifacts (\textit{artf}), (2) background (\textit{bckg}), (3) generalized periodic epileptiform discharge (\textit{gped}), (4) periodic lateralized epileptiform discharge (\textit{pled}), and (5) spike and slow wave (\textit{spsw}). The TUSZ contains EEG signals with manually annotated data for seizure events.

\quad The MMIDB EEG dataset consists of data from 109 participants who are performing or imagining specific motor tasks; our main interest is the moments when subjects either close or imagine closing their left or right fist following a visual cue. We are excluding participants S088, S090, S092, and S100 due to missing data, resulting in 105 participants.

\quad In the construction of brain anatomy concepts, it is imperative to obtain an extensive collection of resting-state EEG data. Due to the limited availability of public datasets with the requisite size and reliability, we utilized The TUH EEG Corpus and source localization to develop a dedicated anatomically labeled resting-state dataset. A set of predefined criteria were employed, including the number of EEG channels, minimum duration, minimum sampling frequency, scaling, and the exclusion of extreme values, which led to the elimination of approximately 90\% of the initial EEG recordings. Following this, a manual examination of a part of the remaining data was performed, ultimately yielding 200 human-verified resting-state EEG recordings, corresponding to an aggregate of about 70 hours of EEG data.

\quad In the process of downstream fine-tuning and concept formation, we employ 19 EEG channels, namely $\textit{Fp1}$, $\textit{Fp2}$, $\textit{F7}$, $\textit{F3}$, $\textit{Fz}$, $\textit{F4}$, $\textit{F8}$, $\textit{T7}$, $\textit{C3}$, $\textit{Cz}$, $\textit{C4}$, $\textit{T8}$, $\textit{T5}$, $\textit{P3}$, $\textit{Pz}$, $\textit{P4}$, $\textit{T6}$, $\textit{O1}$, and $\textit{O2}$ (see the MNE documentation \cite{doi:10.1098/rsta.2011.0081} for more information). These channels originate from the initial pre-training of BENDR using The TUH EEG Corpus. In instances where the datasets lack these channels, we establish the following mapping: $T3 \mapsto T7$, $T4 \mapsto T8$, $P7 \mapsto T5$, and $P8 \mapsto T6$. We also resample the corresponding EEG data to a 256 Hz sampling frequency and apply a high-pass FIRWIN filter with a 0.1 Hz cutoff, a low-pass FIRWIN filter with a 100.0 Hz cutoff, and a 60 Hz FIRWIN notch filter to eliminate powerline noise. In situations where preprocessing cannot be performed, the EEG is excluded. Finally, we scale each trial to the range $[-1, 1]$ and append a relative amplitude channel, see \cite{BENDR}, resulting in a total of 20 channels.


\subsection{Training}
Pre-training of BENDR is based on the large set of unlabelled EEG data from The TUH EEG Corpus. The pre-training procedure is largely based on \verb|wav2vec 2.0| and involves two main stages: The convolutional stage and the transformer stage. The convolutional stage generates a sequence of representations (BENDRs) that summarize the original input. This sequence is then fed into the transformer stage, which adjusts its output to be most similar to the encoded representation at each position. The layers affected during pre-training are the feature encoder and the transformer. Kostas et al.\ \cite{BENDR} kindly made the pre-trained weights of the encoder and contextualizer publicly available, and this is the model that we have employed here.

\quad The LHB model architecture described in Figure \ref{fig:linear_head_bendr} is used for downstream fine-tuning. We aim to optimize the model for two distinct binary classification objectives. First, the model is fine-tuned for the differentiation between \emph{seizure} and \emph{non-seizure} events, using the TUSZ Corpus with 60-second window segments. The hyperparameters are determined using Bayesian optimization to maximize the validation $F_1$-score. The fine-tuning employs a batch size of 80, a learning rate of $1 \times 10^{-4}$, and $30$ epochs. This results in a model with a balanced accuracy of $0.73 \pm 0.07$. 

\quad In our second fine-tuning example, the model is adapted for the differentiation between \emph{Left Fist Movement} versus \emph{Right Fist Movement}, using the MMIDB EEG Dataset with 4-second window segments. We are using both the imaginary and performed task data from the 105 participants. We train the model for $7$ epochs with a batch size of $4$ and a learning rate of $1 \times 10^{-5}$. The hyperparameters were chosen based on the best validation balanced accuracy from leave-one-subject-out cross-validation where the model was trained for 50 epochs and the best model was retained. The specific hyperparameter configuration aligns with the optimal hyperparameters found by the original authors \cite{BENDR} and we find a similar balanced accuracy of $0.83 \pm 0.02$.

\subsection{Constructing Concepts}
\label{subsec:explanatory_concepts}

To construct human-aligned explanatory EEG concepts, a number of initial investigations were conducted. The data processing involved follows the methodology previously mentioned. In this section, we provide a general pipeline overview and discuss several choices made throughout the process.

\vspace{0.5em}
{\bf Concepts from Labeled EEG Data}: Using the labeled EEG data from the TUAR and TUEV Corpus and the MMIDB EEG Dataset, we create concepts representing activities within specific time windows. Each annotated segment of the EEG data is divided into windows of predetermined length and assigned the corresponding label.

\quad In the TUEV Corpus, we define concepts for the spike/short wave (\textit{spsw}), periodic lateralized epileptic discharge (\textit{pled}), general period epileptic discharge (\textit{gped}), technical artifact (\textit{artf}), and background (bckg) with 60-second windows. This approach aligns with the length of the \textit{seizure} classifier.

\quad Lastly, we examine the eye movement (\textit{eyem}) from the TUAR Corpus and \textit{Left Fist Movement} and \textit{Right Fist Movement} from the MMIDB EEG Dataset, both using 4-second windows. These different-sized windows then constitute examples of concepts defined based on their labels.

\vspace{0.5em}
{\bf Anatomical Concepts from Unlabeled EEG Data}:
The objective is to identify concepts representing specific frequency bands within distinct areas of the cortex, e.g. \emph{alpha activity in pre-motor cortex} or \emph{gamma activity in early visual cortex}. To obtain a non-task-specific representation of each cortical area, we utilize resting-state EEG data, as it spontaneously generates activity throughout the cortex. For this purpose, we use a subset of The TUH EEG Corpus, as described above.

\quad To define anatomical concepts, EEG data is segmented into 4-second windows, with the first and last 5 seconds of each sequence  excluded to minimize artifact contamination. The data is then divided into five frequency bands with a FIRWIN bandpass filter: \emph{delta} (0.5-4Hz), \emph{theta} (4-8Hz), \emph{alpha} (8-12Hz), \emph{beta} (12-30Hz), and \emph{gamma} (30-70Hz). The inverse operator for the forward model is computed using eLORETA \cite{doi:10.1098/rsta.2011.0081} via the MNE Python library. Since the spatial resolution is not critical, minimal regularization of $1 \times 10^{-4}$ is applied.

\quad Using the combined version of the multi-modal parcellation of the human cerebral cortex, HCPMMP1 \cite{f8095709e11547daa07262682e1545f2} and the inverse operator, the average power of electrical activity in 23 cortical areas for each hemisphere is determined.

\quad Our interest lies in cortical areas exhibiting the greatest deviation from typical activity within a specific frequency band. However, cortical areas are not equidistant from the scalp or consistent in baseline activity across bands. To normalize for these differences in the distribution of cortical activity, we compute the mean and standard deviation of the power in each cortical area for each frequency band on an EEG session level, which will be employed in various ways. We call these the baseline mean and the baseline standard deviation.

\quad We explore possible approaches to how the baseline means and standard deviation for each EEG session could be used to normalize the power of 4-second windows within that session. The options include dividing by the baseline standard deviation to account for scalp source variation, subtracting or dividing by the baseline mean to identify the cortical area with the greatest deviation, taking the absolute difference or not, and selecting a single cortical area across all frequency bands or only within a specific band.

\quad Identifying a single frequency and cortical area for each 4-second window of EEG data is a challenging task without prior work to guide the process, and each method presents its own limitations. We specifically look for \emph{alpha} desynchronization in the cerebral cortex during imagined or actual movement and closed or open eyes in the MMIDB EEG dataset, i.e., that \emph{alpha} activity in cortical areas decreases when activated.
Using a paired t-test to examine the presence of lateralization in cortical activities for different methods, we found that the preferred approach is to choose the area which maximizes the absolute difference between the given time window's power and the baseline mean, divided by the baseline standard deviation, only within specific frequency bands.

\vspace{0.5em}
{\bf Random Concepts:} Construction of CAVs calls for data examples that are considered random with respect to the concept of interest. In all experiments, random concepts consisting of 4-second or 60-second windows were randomly sampled from resting-state data obtained from the subset of the TUH EEG Corpus and unannotated sections of the TUAR dataset. 

 \subsection{Experiments} \label{subsec:experiments}
We investigate two approaches for defining explanatory concepts in EEG data. The TCAV method is then employed to evaluate whether the LHB model uses specifically defined human-aligned concepts of EEG data. For all concepts, the resulting activation vectors for all five bottlenecks in the LHB model architecture are examined to determine if they significantly align with the latent representations of class data in the model. We conduct the following experiments:
\begin{figure}[b!]
    \centering
    \vspace{-11pt}    \centerline{\includegraphics[width=\linewidth]{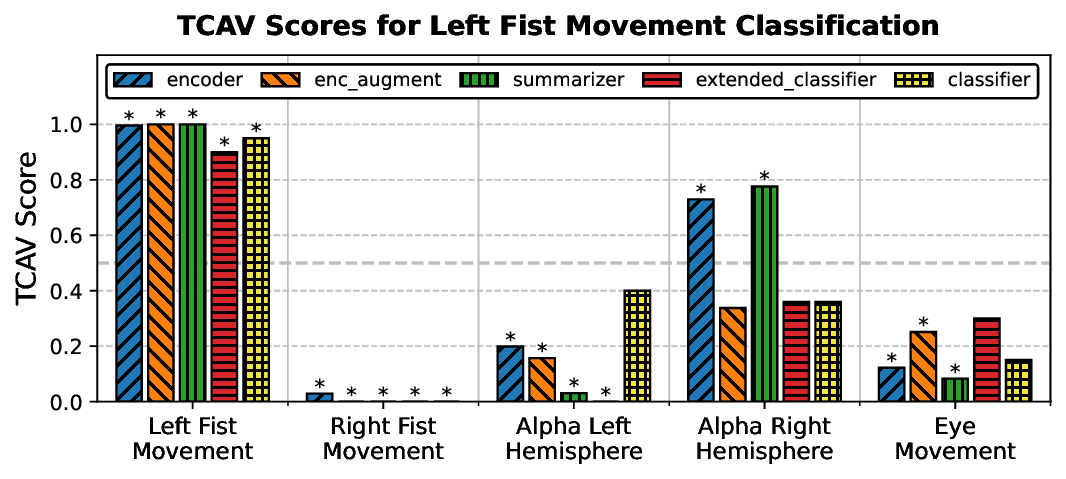}}
    \vspace{-6pt}
    \caption{Sanity checks for applying the TCAV method to EEG data and the bottlenecks of the LHB model. The figure presents the results of TCAV for the \textit{Left Fist Movement} class in a binary classification task using the MMIDB EEG dataset. From right to left, concepts are defined as follows: (1) \textit{Left Fist Movement} and (2) \textit{Right Fist Movement} class data, maximal mean activity in the alpha frequency band for (3) \textit{Left Hemisphere} and (4) \textit{Right Hemisphere}, respectively, and (5) \textit{Eye Movement} artifacts. Stars indicate either positive (a score above 0.5) or negative (a score below 0.5) statistical significance.} 
    \label{fig:results_sanity_check}
    \vspace{-6pt}
\end{figure}
\begin{enumerate}
\item \textbf{Sanity Checks:} We verify the TCAV method and construction of concepts function as intended through a series of sanity checks when classifying \textit{Left Fist Movement}.
\item \textbf{Event-based Concepts:} We assess whether the LHB model leverages specific EEG events in the classification of \textit{seizure}.
\item \textbf{Anatomy/Frequency-based Concepts:} We investigate if the LHB model employs lateralization in cortical activity in the \emph{alpha} band for classifying \textit{Left Fist Movement}. The chosen cortical areas are based on their relevance to the classification task.
\end{enumerate}
In the experiments, we use the TCAV method with a regularized linear model and stochastic gradient descent (SGD) learning, setting the regularization parameter $\alpha = 0.1$ to learn the decision boundary between explanatory and random concepts. We employ 50 random concepts and a maximum of 40 examples per concept. These parameters were chosen to increase statistical power. The mean TCAV scores for the target concept examples and the random examples are compared using the non-parametric Mann-Whitney U Rank test, as opposed to the t-test used in the original TCAV method, as we observed a clear violation of the normality assumption for the TCAV scores. To mitigate Type I errors, the p-values are corrected for each experiment employing the conservative Bonferroni method, after which we claim significance if the corrected p-value is below $0.05$.

\begin{figure}[t!]
    \centering
    \centerline{\includegraphics[width=\linewidth]{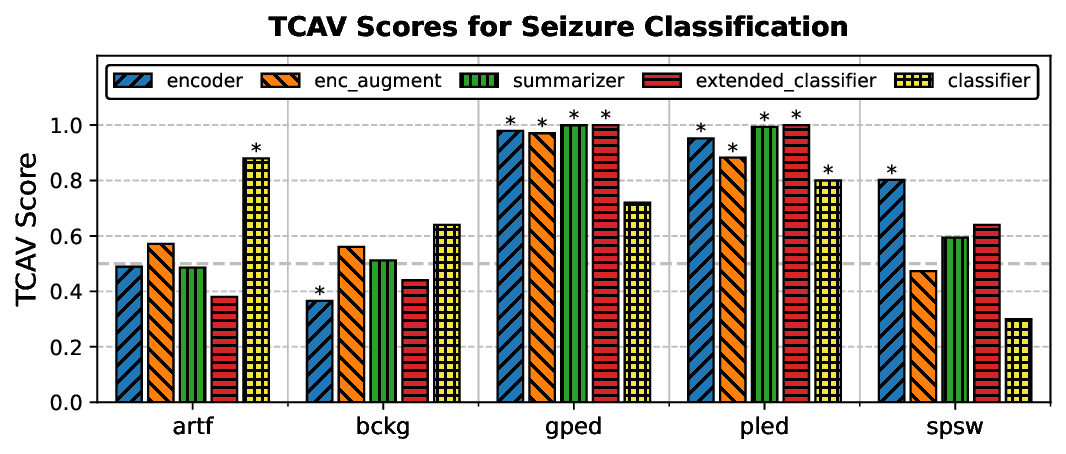}}
    \vspace{-6pt}
    \caption{The results of utilizing TCAV to assess whether event-based EEG labels align with the internal representation of the \textit{seizure} class data in the LHB model at the five bottlenecks are presented. From the right, the concepts are defined as (1) technical artifacts (\textit{artf}), (2) background (\textit{bckg}), (3) generalized periodic epileptic discharge (\textit{gped}), (4) periodic lateralized epileptic discharge (\textit{pled}), and (5) spike and short wave (\textit{spsw}). Stars indicate either positive (a score above 0.5) or negative (a score below 0.5) statistical significance.}
    \label{fig:results_seizure}
    \vspace{-11pt}
\end{figure}
\section{Results} \label{sec:results}
\subsection{Sanity Checks} \label{ssec:sanity_checks}
We first provide evidence that the TCAV method can be applied to explain EEG data and the LHB model. 
In Figure \ref{fig:results_sanity_check}, the high significance of class data as concepts (\textit{Left Fist Movement} with positive evidence and \textit{Right Fist Movement} with negative evidence) confirms this. Furthermore, concepts based on maximal activity in either the left or right hemisphere for the \emph{alpha} frequency band strongly indicate that lateralized cortical activity is detected by several layers in the model, as expected.

\quad Moreover, the negative alignment of a concept based on labeled artifacts with the model representation of motor task data implies that artifacts in EEG data significantly influence classification tasks. We find that \textit{eyem} has a negative impact on the classification of \textit{Left Fist Movement}. Note that this does \textit{not} mean that \textit{eyem} positively affects the opposite class, that is \textit{Right Fist Movement}, as the TCAV Score is specific to the "\textit{Left Fist Movement} dataset". Conversely, \textit{eyem} could negatively affect the classification of both \textit{Left Fist Movement} and \textit{Right Fist Movement}, due to the lower signal-to-noise ratio for classification when artifacts are present.

\subsection{Event-based concepts} \label{ssec:annotated_data_as_concepts}
We next investigate whether fine-tuning the LHB model for seizure classification on the TUSZ dataset and using explanatory concepts defined with labeled data from TUEV aligns with the model's internal representation for data labeled as containing seizures. The target of the investigation is the \textit{seizure} label and we test all bottlenecks in the LHB model. The results of this experiment are shown in Figure \ref{fig:results_seizure}.

\quad When compared to EEG data labeled as containing seizures, the epilepsy-related concepts \textit{pled}, which is present in certain brain areas, and \textit{gped}, which is present in most of the brain, exhibit high and positive evidence in nearly all bottlenecks. This observation aligns with existing literature that associates epileptiform discharges with seizures \cite{gajic2015detection}, and it is expected that the LHB model will use these properties for classification. The \textit{spsw} concept also demonstrates significant positive evidence in the \textit{encoder} bottleneck but not in the further downstream bottlenecks. Similarly, the \textit{bckg} concept shows negative evidence in the \textit{encoder} bottleneck but not in the further downstream bottlenecks. It is interesting that these concepts only come to be significant in the initial bottleneck.
 A possible explanation is that the technical artifacts \textit{artf} and \textit{bckg} are not significant for the classification, but BENDR effectively identifies seizure-related concepts and filters out noise. The results also suggest that the model's \textit{classifier} and \textit{extended classifier} can be further optimized, as \textit{artf} is near-significant level in these bottlenecks and, as a result, the noise has not been completely removed. In conclusion, these examples indicate that concept-based explainability can provide valuable model design information.

\subsection{Anatomy/Frequency-Based Concepts} \label{ssec:artefacts_as_concepts}
\begin{figure*}[h!]
    \centering
    \includegraphics[width=\textwidth]{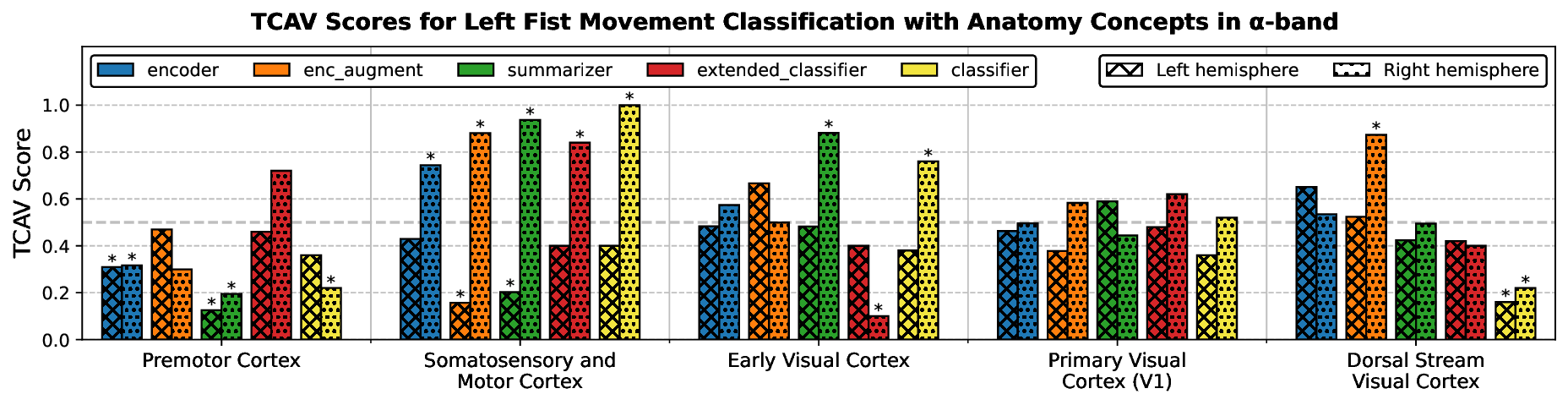}
    \vspace{-20pt}
    \caption{Using TCAV, we analyzed the alignment between anatomical concepts in the \emph{alpha} band and the internal representation of the \textit{Left Fist Movement} class in the LHB model at five bottlenecks. The visualization of five pairs of concepts focused on five cortical areas, located in both the left and right hemispheres, that were deemed most relevant for the classification task. The chosen concepts had a higher deviation in the \emph{alpha} band. Stars indicate either positive (a score above 0.5) or negative (a score below 0.5) statistical significance. Our analysis reveals significant lateralization in the\textit{ Somatosensory and Motor Cortex }across all five bottlenecks. Additionally, we observe that the \textit{Primary Visual Cortex (V1)} was insignificant for both hemispheres in all bottlenecks.}
    \label{fig:results_lateralization}
    \vspace{-11pt}
\end{figure*}
We have demonstrated that labeled EEG data can generate human-aligned concepts, which are integrated into the LHB model for seizure classification. This comes quite naturally as labeled 
data is labeled by humans and tend to align with human-relatable concepts.
We then present evidence that defining explanatory concepts based on cortical activity in frequency bands may uncover patterns corresponding to the model's internal representations.

\quad In particular, for a motor classification task using the MMIDB EEG dataset and targeting the \textit{Left Fist Movement} class, we show that cortical activity in the \emph{alpha} band aligns with the model's internal representation. In Figure \ref{fig:results_lateralization}, we find that the CAV for \textit{Somatosensory and Motor Cortex} in the right hemisphere positively aligns with the activations of \textit{Left Fist Movement} class data across all bottlenecks in the model. The mean TCAV scores are also consistently positively significant. At the same time, the TCAV scores for the same cortical area in the \textit{Left Hemisphere} are either negatively significant or insignificant. These results strongly suggest that the model's internal representation incorporates lateralization, reflecting the fact that one hemisphere exhibits more electrical activity than the other. It is noteworthy that lateralization is most significant in the \textit{Encoding Augment} and \textit{Summarizer} bottlenecks, indicating that it is captured early in the network.

\quad Additionally, we observe that the \textit{Primary Visual Cortex (V1)} areas do not exhibit lateralization, and their TCAV scores are insignificant across all bottlenecks and for both hemispheres. This further supports the conclusion that the LHB model utilizes specific cortical areas in its classification rather than all areas indiscriminately.

\quad While no apparent lateralization is present in the \textit{Premotor Cortex}, this part of the cortex is negatively significant in the \textit{Encoder} and \textit{Summarizer} bottlenecks for both the left and right hemispheres. A possible explanation is that the instances we examine involve participants \textit{performing} movements; therefore, there may not necessarily be relevant activity in the \textit{Premotor Cortex}, which is primarily involved in movement planning \cite{gallego2022going}.

\quad Lastly, we observe significance in the \textit{Classifier} bottleneck for \textit{Early Visual Cortex} and \textit{Dorsal Stream Visual Cortex}. We note that the movement is activated by a visual cue; however, further experiments would be required to fully clarify the effect.
\section{Conclusion} \label{sec:conclusion}
Concept-based explainability has proven to be valuable in various domains, such as image classification and natural language understanding, where concepts are naturally defined using labeled data. In this study, we have explored the definition of concepts for EEG models for the first time. We presented two new workflows for concept-based explainability within the TCAV framework for EEG data. First, we adopted an approach akin to the original work of Kim et al.\ \cite{kim2018interpretability}, in which concepts are derived from labeled data. In this case, we utilized various annotated EEG databases, e.g., data from the Temple University Hospital EEG database. The second workflow is based on the source location of resting-state EEG data also from the Temple University Hospital database. This enables us to generate datasets for TCAV derived from anatomical  brain areas and for specific frequency bands, e.g., the \emph{alpha} band.
We demonstrated a proof of concept through several "sanity check" experiments to verify expected responses in elementary EEG settings, such as EEG lateralization during left- or right-hand movement. Lastly, we examined two practical applications: A case study involving seizure prediction, where TCAV reveals the role of fundamental spike patterns, and a brain-computer interface case, hinting at how the TCAV method can assist in debugging and offer valuable insights into classifier design for EEG data.

\bibliographystyle{IEEEbib}
\bibliography{main}

\begin{thebibliography}{10}

\bibitem{BENDR}
Demetres Kostas, Stéphane Aroca-Ouellette, et~al.,
\newblock ``Bendr: Using transformers and a contrastive self-supervised
  learning task to learn from massive amounts of eeg data,''
\newblock {\em Frontiers in Human Neuroscience}, vol. 15, 2021.

\bibitem{kim2018interpretability}
Been Kim, Martin Wattenberg, et~al.,
\newblock ``Interpretability beyond feature attribution: Quantitative testing
  with concept activation vectors (tcav),'' 2018.

\bibitem{wave2vec}
Alexei Baevski, Henry Zhou, et~al.,
\newblock ``wav2vec 2.0: {A} framework for self-supervised learning of speech
  representations,''
\newblock {\em CoRR}, vol. abs/2006.11477, 2020.

\bibitem{vaswani2017attention}
Ashish Vaswani, Noam Shazeer, et~al.,
\newblock ``Attention is all you need,'' 2017.

\bibitem{BERT}
Jacob Devlin, Ming-Wei Chang, et~al.,
\newblock ``{BERT}: Pre-training of deep bidirectional transformers for
  language understanding,''
\newblock Minneapolis, Minnesota, June 2019, pp. 4171--4186, Association for
  Computational Linguistics.

\bibitem{doi:10.1098/rsta.2011.0081}
Roberto~D. Pascual-Marqui, Dietrich Lehmann, et~al.,
\newblock ``Assessing interactions in the brain with exact low-resolution
  electromagnetic tomography,''
\newblock {\em Philosophical Transactions of the Royal Society A: Mathematical,
  Physical and Engineering Sciences}, vol. 369, no. 1952, pp. 3768--3784, 2011.

\bibitem{tuheeg}
Amir Harati, Silvia Lopez, et~al.,
\newblock ``The tuh eeg corpus: A big data resource for automated eeg
  interpretation,''
\newblock 12 2014.

\bibitem{tuhz}
Vinit Shah, Eva von Weltin, et~al.,
\newblock ``The temple university hospital seizure detection corpus,''
\newblock {\em Frontiers in Neuroinformatics}, vol. 12, 2018.

\bibitem{mmidb}
Gerwin Schalk, Dennis~J. McFarland, et~al.,
\newblock ``Bci2000: a general-purpose brain-computer interface (bci) system,''
\newblock {\em IEEE Transactions on Biomedical Engineering}, vol. 51, no. 6,
  pp. 1034--1043, 2004.

\bibitem{f8095709e11547daa07262682e1545f2}
Matthew~F. Glasser, Timothy~S. Coalson, et~al.,
\newblock ``A multi-modal parcellation of human cerebral cortex,''
\newblock {\em Nature}, vol. 536, no. 7615, pp. 171--178, Aug. 2016.

\bibitem{gajic2015detection}
Dragoljub Gajic, Zeljko Djurovic, et~al.,
\newblock ``Detection of epileptiform activity in eeg signals based on
  time-frequency and non-linear analysis,''
\newblock {\em Frontiers in computational neuroscience}, vol. 9, pp. 38, 2015.

\bibitem{gallego2022going}
Juan~A. Gallego, Tamar~R. Makin, et~al.,
\newblock ``Going beyond primary motor cortex to improve brain--computer
  interfaces,''
\newblock {\em Trends in neurosciences}, 2022.

\end{thebibliography}

\end{document}